%% file: main.tex
\documentclass[10pt,twocolumn,letterpaper]{article}

\usepackage{wacv}              

\usepackage{graphicx}
\usepackage{amsmath}
\usepackage{amssymb}
\usepackage{booktabs}
\usepackage{mathabx}
\usepackage{multirow}
\usepackage{verbatim}
\usepackage[numbers,sort,compress]{natbib}
\usepackage[dvipsnames]{xcolor}
\usepackage[normalem]{ulem}
\usepackage[accsupp]{axessibility}  

\usepackage[pagebackref,breaklinks,colorlinks]{hyperref}

\usepackage[capitalize]{cleveref}
\crefname{section}{Sec.}{Secs.}
\Crefname{section}{Section}{Sections}
\Crefname{table}{Table}{Tables}
\crefname{table}{Tab.}{Tabs.}

\def\wacvPaperID{62} 
\def\confName{WACV}
\def\confYear{2024}

\begin{document}

\title{OTAS: Unsupervised Boundary Detection for Object-Centric\\
Temporal Action Segmentation}


\twocolumn[{%
\renewcommand\twocolumn[1][]{#1}%
\maketitle
\vspace{-20mm}
\begin{center}
\large{
Yuerong Li\,\textsuperscript{1},\;
Zhengrong Xue\,\textsuperscript{2,3,4},\;
Huazhe Xu\,\textsuperscript{2,3,4}\\
\vspace{2mm}
\textsuperscript{1}\,Zhejiang University \;
\textsuperscript{2}\,Tsinghua University \; \textsuperscript{3}\,Shanghai Qi Zhi Institute \; \textsuperscript{4}\,Shanghai AI Lab \\
}
\vspace{2mm}
\normalsize\url{https://github.com/yl596/OTAS}
\vspace{6mm}
\end{center}
}]

\input{00abstract}
\input{01intro}
\input{02relatedwork}

\input{03method}
\input{04experiment}

\input{05conclusion}

{\small
\bibliographystyle{ieee_fullname}
\bibliography{main}
}

\end{document}

%% file: 00abstract.tex
\begin{abstract}
Temporal action segmentation is typically achieved by discovering the dramatic variances in global visual descriptors.
In this paper, we explore the merits of local features by proposing the unsupervised framework of \underline{O}bject-centric \underline{T}emporal \underline{A}ction \underline{S}egmentation (OTAS). Broadly speaking, OTAS consists of self-supervised global and local feature extraction modules as well as a boundary selection module that fuses the features and detects salient boundaries for action segmentation. 
As a second contribution, we discuss the pros and cons of existing frame-level and boundary-level evaluation metrics. Through extensive experiments, we find OTAS is superior to the previous state-of-the-art method by $41\%$ on average in terms of our recommended F1 score. Surprisingly, OTAS even outperforms the ground-truth human annotations in the user study. Moreover, OTAS is efficient enough to allow real-time inference.
\end{abstract}

%% file: 01intro.tex
\section{Introduction}

\begin{figure}[t]
\centering
\includegraphics[width=0.46\textwidth]{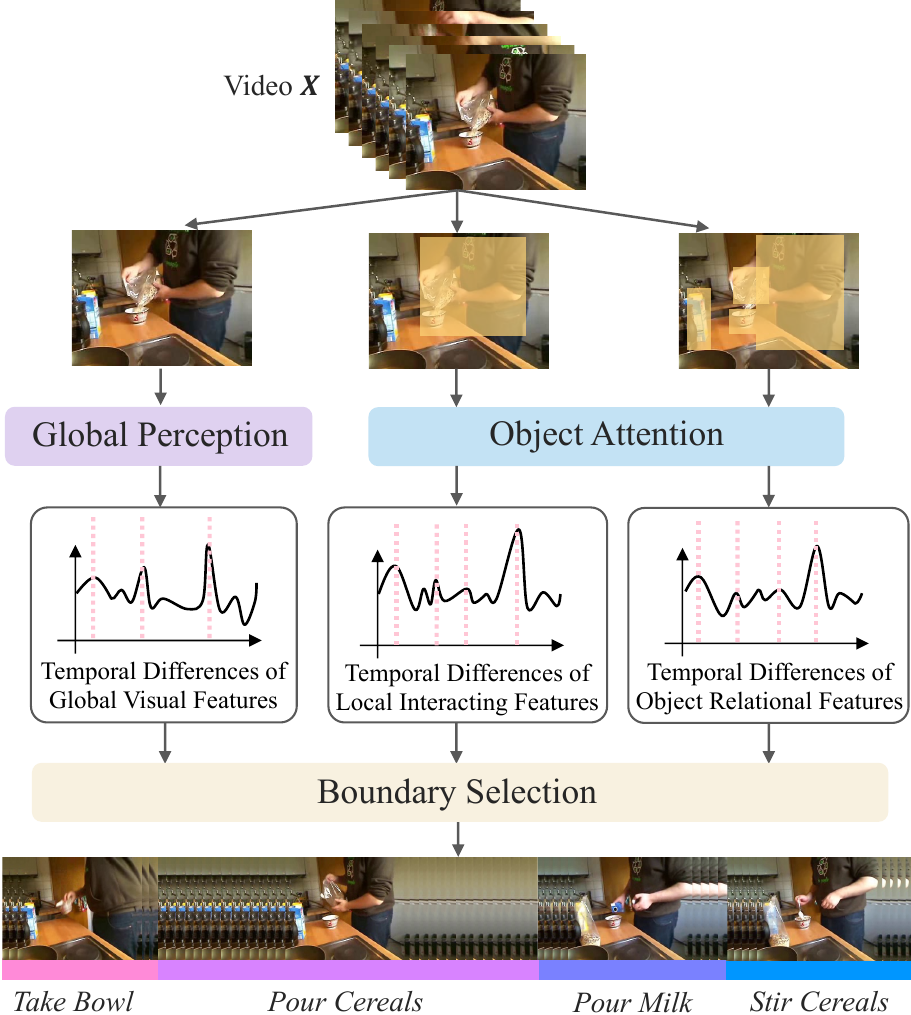}
\vspace{0.9mm}
\caption{\textbf{The framework of Object-centric Temporal Action Segmentation (OTAS).} Given some untrimmed sequences of frames, OTAS learns global visual features, local interacting features, and object relational features via the self-supervised global perception and object attention modules. Afterwards, the boundary selection module fuses the generated global and local features to detect the salient boundaries for action segmentation.}
\label{fig:pipeline}
\end{figure}

Temporal action segmentation~\cite{ding2022temporal} aims to label every frame in an untrimmed video with action tags. With the emergence of large-scale instructional video datasets~\cite{kuehne2014language,stein2013combining,alayrac2016unsupervised}, various learning-based frameworks~\cite{kuehne2014language,richard2017weakly,chen2020action} tackle action segmentation under supervision. However, densely-annotated video datasets are often criticized to be excessively expensive; even worse, human-decided action boundaries are subjective, leading to non-negligible biases.

In the more tempting fully unsupervised scenario where the action types are unavailable, a set of predicted action boundaries could sufficiently infer the untagged temporal segmentation. Thus, there appear two branches of works: one branch~\cite{kukleva2019unsupervised,li2021action,sarfraz2021temporally} learns to group similar frames and transforms the segmentation problem into a clustering process; the other branch~\cite{Aakur_2019_CVPR,wang2021coseg,du2022fast} learns to detect the boundaries that indicate the most salient variations among the frames and generates segmentation afterwards.

\begin{figure}[t]
\centering
\includegraphics[width=0.46\textwidth]{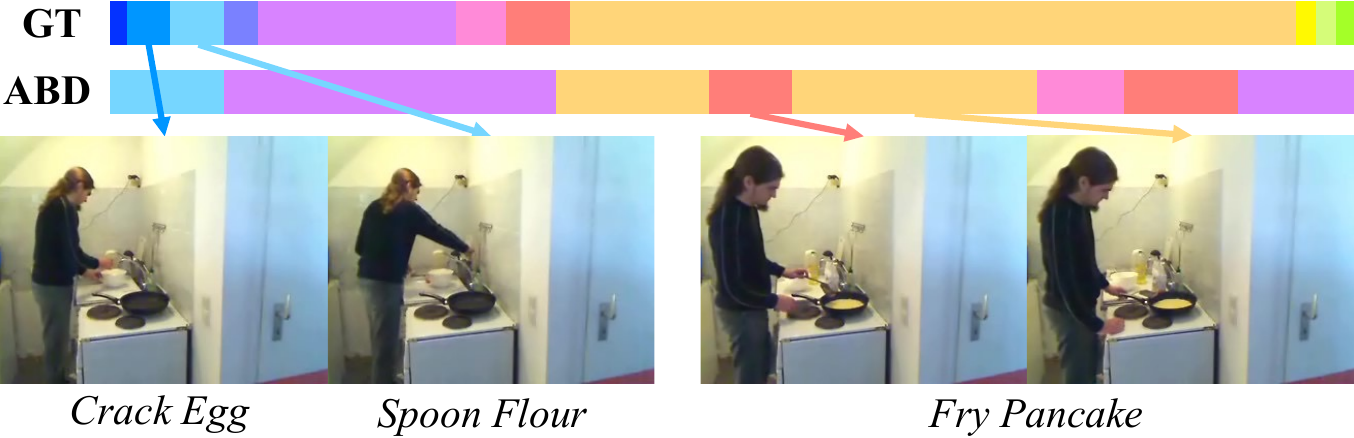}
\caption{\textbf{Examples of the common failure modes of the SOTA ABD~\cite{du2022fast} method.} ABD fails to distinguish the fine discrepancy between \textit{Crack Egg} and \textit{Spoon Flour} as the actions occur in the far end of the camera perspective. Meanwhile, it over-segments \textit{Fry Pancake} due to the whole-body motions of the human subject.}
\label{fig:motivation}
\end{figure}

While the current research hotspots for unsupervised action segmentation are mainly the elaborately designed clustering or boundary selection techniques, a crucial yet less studied component in this field lies in the very early stage of the pipeline, \ie, the feature extraction module. Conceptually, we highlight the importance of the extracted features in that they fundamentally determine the criteria for grouping or distinguishing the adjacent frames. By analyzing the failure modes of existing state-of-the-art methods, we further argue that general-purpose global extractors, either plain CNNs~\cite{Aakur_2019_CVPR,wang2021coseg} or learning-free descriptors~\cite{sarfraz2021temporally,kukleva2019unsupervised,du2022fast} such as IDT~\cite{wang2013action}, do not necessarily best fit the task of temporal action segmentation. Specifically, existing extractors pay equal attention to all the details in the video clip. Thus, the subsequent segmentation might be easily interfered with semantically characterless but numerically dramatic variations such as camera perspective shift or arbitrary large-scale movements of the human subject. In Figure~\ref{fig:motivation}, we provide concrete examples to shed light on our observations.

Inspired by the cognitive grounds~\cite{zacks2007event,biederman1987recognition} that humans rely on partial components to segment complex activities, we attempt to alleviate the incompetence of global visual descriptors by exploring the merits of object-centric local features. 
To enhance the instantiation of this idea, we propose the framework of Object-centric Temporal Action Segmentation (OTAS),
as illustrated in Figure~\ref{fig:pipeline}. More specifically, the devised framework includes three major modules: a) a self-supervised global perception module that learns global visual features; b) a self-supervised object attention module that captures local interacting features and inter-object relational features; c) a boundary selection module that fuses the features and selects salient boundaries.

Another highlight of this paper is a discussion on the evaluation metrics, where we point out popular frame-level metrics such as Mean of Frames (MoF) are likely to be manipulated by the dominant pattern in the sequence, while the recognized acceptance thresholds of boundary-level F1 scores are far too coarse. Therefore, we appeal to the community for more thoughtful designs of metrics and in the meantime recommend a revised F1 score. 

In terms of the recommended metrics, our OTAS outperforms the previous state-of-the-art approach~\cite{du2022fast} by up to $\textbf{41\%}$ on average. In terms of MoF, OTAS is superior to all the baselines on the Breakfast~\cite{kuehne2014language} and 50Salads~\cite{stein2013combining} datasets.
As supplements to the quantitative results, we conduct a user study and a qualitative case study, which cross-validate the competitive performance of our approach and the rationality of our claim on evaluation metrics.
Furthermore, by directly consuming raw video inputs, OTAS is found to be efficient enough to perform real-time inference.

\begin{figure*}[ht]
\centering
\includegraphics[width=0.98\textwidth]{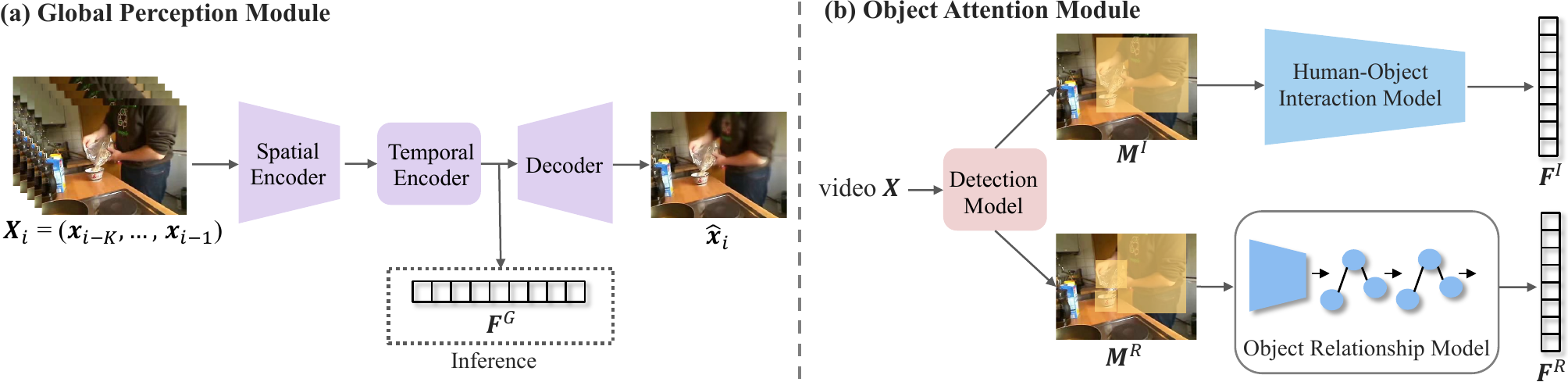}
\vspace{0.15mm}
\caption{\textbf{The pipeline of feature extraction. (a) Global perception module} takes as input a sequence of $K$ consecutive frames and outputs the \textit{global visual features} $\boldsymbol{F}^G$. \textbf{(b) Object attention module} leverages an off-the-shelf detection model to obtain both interactive region and object masks. The interactive regions pass through a human-object interaction model to produce \textit{local interacting features} $\boldsymbol{F}^I$. Meanwhile, the object masks are consumed by an object relationship model to generate \textit{object relational features} $\boldsymbol{F}^R$. Both modules depend on frame prediction to validate self-supervised training.}
\label{model}
\end{figure*}

%% file: 02relatedwork.tex
\section{Related Works}
\subsection{Supervised Action Segmentation}
Temporal action segmentation~\cite{ding2022temporal} has been thoroughly studied under fully supervised settings~\citep{kuehne2014language,kuehne2016end,chen2020action,farha2019ms,zhou2018towards,tang2020comprehensive}, where various kinds of recurrent models are leveraged for the sequential prediction of the action labels.

As the annotation process is expensive 
, researchers also investigate methods that harness
weaker supervision such as narrations or subtitles 
~\citep{bojanowski2015weakly,sener2015unsupervised,alayrac2016unsupervised,malmaud2015s,miech2019howto100m,miech2020end,shen2021learning}. While the language guidance does provide some assistance, the temporal misalignment is 
inevitable since the spoken words are often ahead of the actual actions. Another line of weakly supervised works relieves the need for language guidance, where only the action types and their orders are provided~\citep{richard2017weakly,kuehne2017weakly,ding2018weakly,richard2018neuralnetwork,li2019weakly,wang2021end}. Furthermore, there are works relying on different weak supervision options such as an unordered set of actions~\cite{richard2018action} or timestamps~\cite{li2021temporal}.

\subsection{Unsupervised Action Segmentation}
Due to expensive labor and inevitable biases brought by human annotations, unsupervised methods for action segmentation are increasingly popular. Unaware of the action types, unsupervised methods divide the video clip into neutral segments assigned with no specific classes and then depend on the Hungarian matching algorithm~\cite{kuhn1955hungarian} to establish the correspondence between unlabeled predictions and ground-truth action classes.
Generally, unsupervised methods can be categorized into clustering-based ones that group similar frames as well as boundary-based ones that detect salient variations.

\noindent\textbf{Clustering-based methods.}
Many of the clustering-based works either assume that the 
videos contain the same activity~\citep{sener2018unsupervised,kukleva2019unsupervised,elhamifar2019unsupervised,li2021action,vidalmata2021joint,wang2021unsupervised,swetha2021unsupervised,wang2022sscap} or target at multi-activity collections but 
hold known activity labels~\citep{elhamifar2020self,ding2021temporal,kumar2021unsupervised}. A fully unsupervised pipeline is pioneered by \citet{sener2018unsupervised}, who develop an iterative approach that alternates between a discriminative visual feature model and a generative temporal model.
Meanwhile, clustering unknown activities is first explored by CTE~\cite{kukleva2019unsupervised}, where continuous temporal embeddings are learned to simultaneously capture time dependencies and visual representations.

Recently, more advanced techniques such as action shuffle alternating~\cite{li2021action} and temporally-weighted hierarchical clustering (TW-FINCH)~\cite{sarfraz2021temporally} are developed to boost the segmentation performance. Nevertheless, the fact that clustering is 
an iterative process makes it computationally inefficient, hindering its deployment in many actual scenarios.

\noindent\textbf{Boundary-based methods.}
In comparison with clustering-based counterparts that densely annotate all of the frames, boundary-based approaches are more efficient thanks to a sparse and one-pass boundary discovery process. Inspired by cognitive psychology, LSTM+AL~\cite{Aakur_2019_CVPR} is one of the first unsupervised works that propose to detect the boundaries by analyzing the curve of errors in a self-supervised frame prediction procedure. Likewise, CoSeg~\cite{wang2021coseg} generate temporal features via contrastive learning and detect the boundaries by leveraging Transformer~\cite{vaswani2017attention} models. Lately, ABD presented by \citet{du2022fast} proposes to detect the boundaries with learning-free offline features and apply a clustering algorithm for refinement.

Despite their remarkable performance, all of the existing methods only consider global visual features, making them vulnerable to semantically trivial but numerically predominant action variations. In contrast to the previous works, we emphasize on the feature extraction module, arguing the local interactive and relational features are also indispensable to the judgment of the ongoing actions.

\subsection{Generic Event Boundary Detection}
A closely related topic to temporal action segmentation is the newly proposed problem known as generic event boundary detection (GEBD)~\citep{shou2021generic,kang2021uboco,tang2022progressive}, where changes in perspectives, color/brightness, subjects, {\it etc}, are all considered as target boundaries. Unlike GEBD, we aim at the more prevailing objective
of action segmentation, \ie, the unsupervised detection of the semantically salient transitions from an untrimmed sequence of actions.

%% file: 03method.tex
\section{Method}
We propose OTAS, an unsupervised framework for boundary-based temporal action segmentation that considers not only global visual features but also local object-centric features. In this section, we first present the formulation of action segmentation in Section~\ref{section:formula}. Then, the global perception module for global feature extraction is introduced in Section \ref{section:FPM}, while the object attention module for object feature extraction is introduced in Section \ref{section:OAM}. Finally, the resulting features are fused by the boundary selection module 
described in Section \ref{section:BDM} for final boundaries.

\subsection{Problem Formulation}
\label{section:formula}
We denote a video of unknown activity as a frame sequence $\boldsymbol{X}=\left(\boldsymbol{x}_1, \dots, \boldsymbol{x}_L\right)$ of length $L$, where $\boldsymbol{x}_i$ is the $i$-th RGB frame. Typically, $\boldsymbol{x}_i$ is in the shape of $H \times W \times C$, where $H$, $W$, $C$ are the height, width, and number of channels of the frame, respectively. From the perspective of boundary-based methods, temporal action segmentation is equivalent to determining a boundary set $\mathcal{B}=\left\{B_m\right\}^{M-1}_{m=1}$ where $B_m \in \left\{1,\dots, L-1 \right\}$, which could split $\boldsymbol{X}$ into $M$ semantically atomic actions.

\subsection{Global Perception Module}
\label{section:FPM}
Endorsed by a number of previous works~\cite{Aakur_2019_CVPR,wang2021coseg,du2022fast}, global visual features could play an important role in boundary detection for temporal action segmentation. Following the practice of~\cite{Aakur_2019_CVPR,wang2021coseg}, we develop a global perception module for self-supervised global feature extraction.
As depicted in Figure~\ref{model}(a), the spatial-temporal encoder takes as input a sequence of $K$ consecutive past frames $\boldsymbol{X}_i=\left( \boldsymbol{x}_{i-K}, \dots,  \boldsymbol{x}_{i-1}\right)$, and the decoder outputs the predicted current frame $\widehat{\boldsymbol{x}}_i$.
To optimize the encoder-decoder architecture in an end-to-end manner, Mean Squared Error (MSE) is adopted for loss computation.
The global visual feature is extracted from the backbone layer at inference.

Specifically, ResNet-50~\citep{he2016deep} serves as the spatial encoder to transform a sequence of high-dimensional raw image inputs into the sequence of low-dimensional embeddings:
\begin{equation} 
\boldsymbol{E}_i= \mathrm{SpatEnc}(\boldsymbol{X}_i),
\;
\mathrm{SpatEnc}:\mathbb{R}^{K \times H \times W \times C}\to \mathbb{R}^{K \times D_E} \notag,
\end{equation}
where $\boldsymbol{E}_i=\left( \boldsymbol{e}_{i-K}, \dots,  \boldsymbol{e}_{i-1}\right)$, and $\boldsymbol{e}_t \in \mathbb{R}^{D_E}$.

Next, a cascading Transformer~\citep{vaswani2017attention} architecture is leveraged to model the latent temporal relations underlying the sequence of spatial embeddings:
\begin{equation} 
\boldsymbol{F}_i^G= \mathrm{TempEnc}(\boldsymbol{E}_i),
\;
\mathrm{TempEnc}:\mathbb{R}^{K \times D_E}\to \mathbb{R}^{K \times D_E} \notag,
\end{equation}
where $\boldsymbol{F}_i^G=(\boldsymbol{f}_{i-K}^G, \dots,  \boldsymbol{f}_{i-1}^G)$, and $\boldsymbol{f}_{t}^G\in\mathbb{R}^{D_E}$ is the global visual feature of the $t$-th frame.

To validate self-supervised training, several up-sampling and convolutional layers are stacked to decode the predicted current frame from the global visual feature of the previous frame:
\begin{equation}
\widehat{\boldsymbol{x}}_i=\mathrm{Decoder}(\boldsymbol{f}_{i-1}^G) , \; \mathrm{Decoder}:  \mathbb{R}^{D_E} \to \mathbb{R}^{H \times W \times C} \notag.
\end{equation}

Once trained through the frame prediction process, the network produces distinctive and time-correlated global features from the bottleneck layer, making preparations for boundary selection to be discussed in Section \ref{section:BDM}. More details on the network architectures and the training process can be found in the supplementary materials.


\subsection{Object Attention Module}
\label{section:OAM}
The sole global features might be easily disturbed by semantically characterless but numerically dramatic variations such as arbitrary whole-body movements of the human.
Therefore, as assistance to the global features, we devise an object attention module to capture the interaction among objects and their interaction with humans. As shown in Figure~\ref{model}(b), the object attention module consists of an off-the-shelf detection model for pre-possessing, a human-object interaction model (Section \ref{section:HIM}) for local interacting features, and an object relationship model (Section \ref{section:ORM}) for object-relational features.

\subsubsection{Human-Object Interaction Model}
\label{section:HIM}
To model the interaction between humans and objects, 
we rely on an out-of-domain off-the-shelf detection model~\citep{ren2015faster} from Detectron2~\citep{wu2019detectron2} pre-trained on the COCO dataset~\citep{lin2014microsoft} to provide a mask $\boldsymbol{m}_{t}$ covering both the human and the interacted objects for each frame.
Thus, the sequence of masked input frames $\widebar{\boldsymbol{X}}=(\widebar{\boldsymbol{x}}_{1}, \dots, \widebar{\boldsymbol{x}}_{L})$ that concentrate on human-object interactions can be acquired by:
\begin{equation}
\widebar{\boldsymbol{X}}=\boldsymbol{X} \odot \boldsymbol{M}, \notag
\end{equation}
where $\boldsymbol{M}=(\boldsymbol{m}_{1}, \dots, \boldsymbol{m}_{L})$ is the sequence of frame-wise masks and $\odot$ refers to the element-wise product. 
With the masked frame sequence ready, we pass it through exactly the same encoder as that trained via the global perception module in Section~\ref{section:FPM}. Since the outputs are supposed to contain rich interactive information, they are named as local interacting features $\boldsymbol{F}^I=(\boldsymbol{f}_1^I, \dots, \boldsymbol{f}_L^I), \boldsymbol{F}^I \in \mathbb{R}^{L \times D_E}$, which are also prepared for the boundary selection module.

\subsubsection{Object Relationship Model}
\label{section:ORM}
\begin{figure}[tb]
\centering
\includegraphics[width=0.46\textwidth]{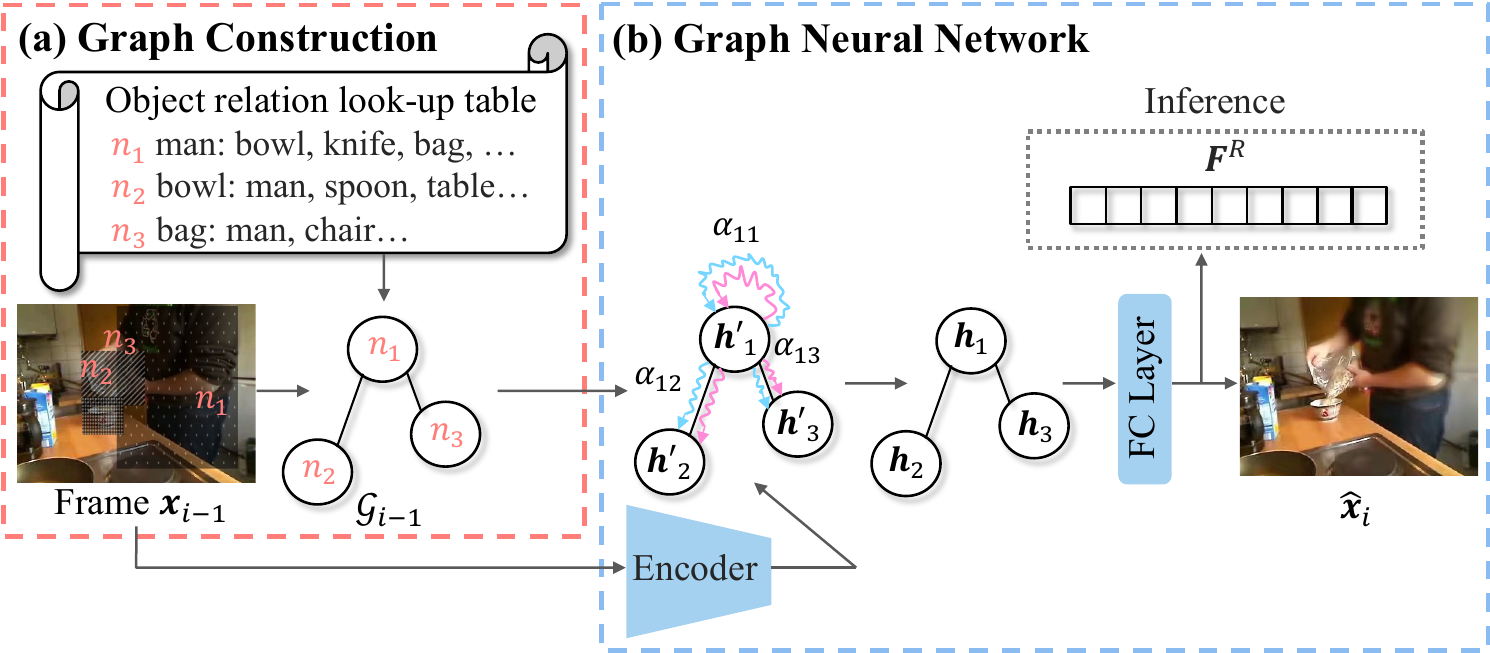}
\caption{\textbf{Object relationship model}. \textbf{(a) Graph construction.} The graph is built upon the object relation look-up table while the object masks are generated from the object detection model. \textbf{(b) Graph Neural Network.} The object masks are consumed by an encoder to initialize each node's representation. Then, multi-head self-attention is used to process the graph information. Finally, a fully connected layer is applied to compute the object relationship features. Here, we omit the decoder at training time for clarity.}
\label{object relationship model}
\end{figure}

Apart from human-object interactions, object-object relations are also helpful indicators for action segmentation. OTAS proposes to model the relationships of the detected object patches using graphs, as shown in Figure~\ref{object relationship model}. More specifically, an object relation look-up table~\citep{yang2018visual} is utilized to construct a relational graph for each frame, connecting semantically reactive and spatially neighboring objects. Afterward, a Graph Neural Network (GNN)~\citep{brody2021attentive} is trained by predicting the future frames from both the image features and the cropped object features. During inference time, latent features of the GNN are extracted, known as the object-relational features.

\noindent\textbf{Object relation look-up table.}
We aim to extract relations among objects that are either similar in semantics or possible for mutual interactions. Hence, we build an object relation look-up table by collecting the relations from the Visual Genome dataset~\citep{krishna2017visual} which concern objects concurrently appearing in the COCO dataset.

\noindent\textbf{Graph construction.}
We construct a graph $\mathcal{G}=\left(\mathcal{V},\mathcal{E}\right)$ of object relations for each frame. The nodes $\mathcal{V}=\left\{v^{(1)},...,v^{(N)}\right\}$ correspond to the $N$ detected objects within the frame.  
The edge $\{v^{(p)},v^{(q)}\}$ between node $v^{(p)}$ and node $v^{(q)}$ exists if and only if a) $v^{(p)}$ and $v^{(q)}$ are semantically related according to the look-up table; and b) the distance between the two masks of the objects is within a threshold $\theta_r$. In Figure~\ref{object relationship model}(a), we illustrate how to construct the graph with the help of the object relation look-up table.

\noindent\textbf{Graph Neural Network.}
To extract the latent inter-object relationship information embedded in the object graph, we train the network by forcing it to predict the current frame $\widehat{\boldsymbol{x}}_{i}$ from the information contained in the past frame $\boldsymbol{x}_{i-1}$.
A visual illustration is shown in Figure~\ref{object relationship model}(b).

For each node $v_i^{(p)} \in \mathcal{V}_i$, 
its initial node representation $\boldsymbol{r}_i^{(p)} \in \mathbb{R}^{H \times W \times 2C}$ is the concatenation of the whole image frame and the masked patch of the $p$-th object:
\begin{equation}
\boldsymbol{r}_i^{(p)}=\boldsymbol{x}_i || (\boldsymbol{x}_i \odot \boldsymbol{m}_i^{(p)}), \notag
\end{equation}
where $\boldsymbol{m}_i^{(p)}$ is the mask of the $p$-th detected object in the $i$-th frame, and $||$ denotes the concatenation operation.

Our detailed implementation of the Graph Neural Network is adapted from the dynamic graph attention variant of Graph Attention Networks (GATs) \citep{velickovic2017graph, brody2021attentive}. Specifically, a self-attention mechanism for node $v_i^{(p)}$ is used to attend over its neighbors $\mathcal{N}_i^{(p)}=\{v_i^{(q)} \in \mathcal{V}_i \mid \{ v_i^{(p)},v_i^{(q)} \} \in \mathcal{E}_i\}$, where a scoring function $s$ computes the attention score of a node $v_i^{(p)}$ and its neighbor $v_i^{(q)}$: 
\begin{align}
 &s
 \left(\boldsymbol{r}_i^{(p)}, \boldsymbol{r}_i^{(q)}\right)  = 
\boldsymbol{a}
^{\top}
	\mathrm{LeakyReLU}
	\left(
		\boldsymbol{W} \cdot (\boldsymbol{r}_i^{(p)} || \boldsymbol{r}_i^{(q)} )
	\right), \nonumber
\end{align}
where $\boldsymbol{a} \in \mathbb{R}^{2D_E}$ and $\boldsymbol{W} \in \mathbb{R}^{2D_E\times 2D_E}$ are learnable parameters. Then, the scoring function is leveraged to calculate the hidden node representation $\boldsymbol{h}_i^{(p)}\in \mathbb{R}^{2D_E}$: 
\begin{align}
    \alpha_i^{pq} &=
	\frac{\mathrm{exp}\left(s\left(\boldsymbol{r}_i^{(p)}, \boldsymbol{r}_i^{(q)}\right)\right)}{\sum\nolimits_{v_i^{(q')}\in\mathcal{N}_i^{(p)}} \mathrm{exp}\left(s\left(\boldsymbol{r}_i^{(p)}, \boldsymbol{r}_i^{(q')}\right)\right)}, \notag \\ 
 	\boldsymbol{h}_i^{(p)} &= \text{ReLU}
	\left(
		\sum\nolimits_{v_i^{(q)}\in\mathcal{N}_i^{(p)}}
		\alpha_i^{pq}
		\boldsymbol{W}\cdot\boldsymbol{r}_i^{(q)}
	\right). \notag
\end{align}
The entire hidden representation $\boldsymbol{H}_i \in \mathbb{R}^{N \times 2D_E}$ of a frame is the concatenation of the hidden representations for all the nodes in $\mathcal{V}_i$. Finally, an additional fully-connected layer consumes $\boldsymbol{H}_i$ and outputs the final object relational feature $\boldsymbol{F}^R_i \in \mathbb{R}^{D_E}$. Besides, at training time, $\boldsymbol{F}^R_i$ is further fed to a decoder similar to that in Section~\ref{section:FPM} for frame prediction so as to enable the self-supervised training process.



\subsection{Boundary Selection Module}
\label{section:BDM}
\begin{figure}[tb]
\centering
\includegraphics[width=0.46\textwidth]{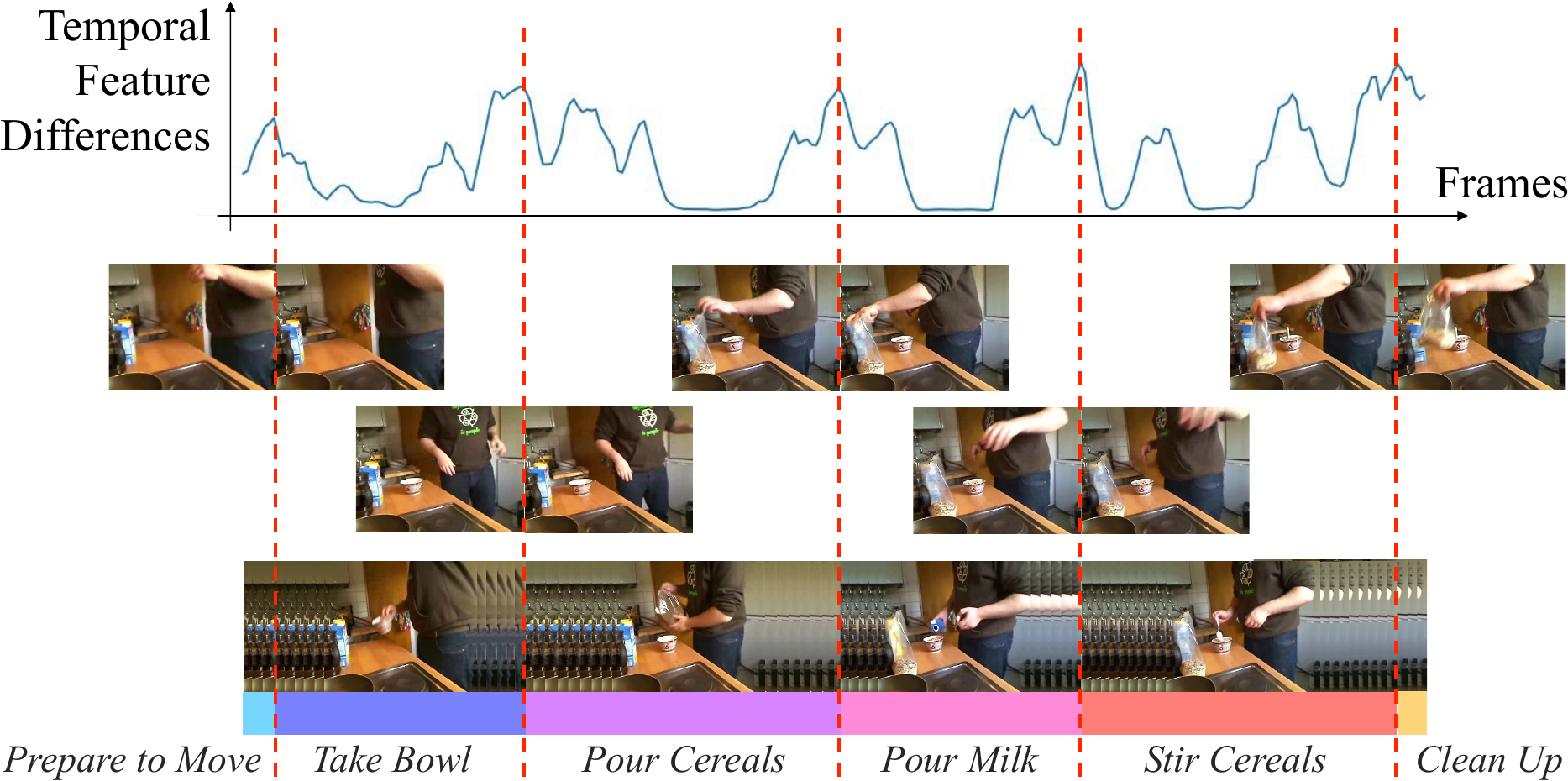}
\vspace{0.1mm}
\caption{\textbf{Boundary selection.} Local maximums of the temporal feature differences are regraded as the boundary candidates, who vote to decide the final boundary predictions.}
\label{Boundary detection }
\end{figure}

Given the access to the global visual features $\boldsymbol{F}^G$, the local interacting features $\boldsymbol{F}^I$, and the object-relational features $\boldsymbol{F}^R$, the boundary selection module is desired to effectively integrate the three features and sensitively capture the salient variations. As shown in Figure~\ref{Boundary detection }, we select the final boundaries from the local maximums of the sequence of temporal feature differences.

For an arbitrary feature $\boldsymbol{f}_i^X$ of the frame $\boldsymbol{x}_i$ taken from the set $\{\boldsymbol{f}_i^G,\boldsymbol{f}_i^I,\boldsymbol{f}_i^R \}$, we define its temporal feature difference $\epsilon_i^X$ as the accumulated squared-$\ell2$ distance between a sequence of $K$ consecutive features prior to $\boldsymbol{x_i}$, \ie, $(\boldsymbol{f}_{i-K}^X, \dots, \boldsymbol{f}_{i-1}^X)$, and another sequence of $K$ consecutive frames starting from $\boldsymbol{x_i}$, \ie, $(\boldsymbol{f}_i^X, \dots, \boldsymbol{f}_{i+K-1}^X)$. 
Then, a frame is considered as a boundary candidate if its temporal feature difference is the local maximum within an interval $\alpha$.
We obtain three sets of boundary candidates from 
the sequence of global feature differences $\epsilon^G$, interacting feature differences $\epsilon^I$, and object-relational feature differences $\epsilon^R$.


Now with the candidates at hand, we design a specialized voting mechanism to determine the final boundary predictions. In broad strokes, we accept two types of boundaries: a) a boundary that is agreed by all of the three candidate sets; or b) a boundary whose temporal feature difference is significantly salient. To instantiate this idea, we calculate a weighted confidence score for all the candidate boundaries: 
\begin{equation}
S=\beta_G \epsilon^G +  \beta_I \epsilon^I+\beta_R \epsilon^R,\notag
\end{equation}
where $\beta_G,\beta_I, \beta_R$ are tunable.
To obtain the boundaries that are unanimously agreed by all the sets, we examine each candidate generated from the global features. If it has neighboring candidates from both other sets within a small time range $\theta_n$, we call these neighbors constituting a boundary cluster. Within each boundary cluster, we select the one with the highest confidence score $S$ as the final prediction. 
For those candidates that are significantly salient in one specific feature but do not have neighbors from other sets, we accept those enjoying the confidence scores that are twice larger than the maximum of the previously selected ones.

%% file: 04experiment.tex
\section{Experiments}
We first introduce the experimental setups in Section~\ref{section:setup}. Next, we call for the community's attention to more reasonable metrics for unsupervised temporal action segmentation in Section~\ref{section:metrics}. Then, we compare the proposed OTAS with competitive baselines in Section~\ref{section:results} in terms of quantitative and qualitative performance, user study, and computational cost. Lastly, we ablate the important factors that contribute to the effectiveness of OTAS in Section~\ref{section:ablation}.

\subsection{Setup}
\label{section:setup}
\subsubsection{Datasets}
\noindent\textbf{Breakfast~\citep{kuehne2014language}} includes 1,712 videos comprising 10 breakfast cooking activities. The duration of the videos varies dramatically from 30 seconds to 7 minutes. Furthermore, the videos may contain occlusions and different viewpoints.

\noindent\textbf{50Salads~\citep{stein2013combining}} includes altogether 4 hours videos of mixing salad, whose average length is 5 minutes.It defines 17 \textit{mid-level} activities and 9 \textit{eval-level} activities.

\noindent\textbf{INRIA~\citep{alayrac2016unsupervised}} includes 150 videos of complex activities not restricted to cooking with an average length of 2 minutes. Specially, it carries up to 83\% ratio of background frames that are extremely diverse in terms of visual appearance.

\subsubsection{Hyper-Parameters}
We down-sample the rates of all the videos to $5\:\mathrm{fps}$ and resize each frame to $256 \times 256 \times 3$. For hyper-parameters, we set $D_E=2048$ as the feature dimension, $K=5\:\mathrm{frames}$ which lasts for one second as the sequence length for frame prediction and boundary selection, $\theta_r=80\:\mathrm{pixels}$ for the distance threshold between two object masks in graph construction, and $\alpha=15\:\mathrm{frames},\;\beta_G=\beta_I=1,\:\beta_R=0.3,\;\theta_n=2\:\mathrm{seconds}$ for the boundary selection module.

More implementation details can be found in the supplementary materials.

\subsubsection{Baselines}
\label{B}
OTAS is compared with a series of competitive baselines including the clustering-based CTE~\citep{kukleva2019unsupervised} and TW-FINCH\citep{sarfraz2021temporally}, and the boundary-based LSTM+AL~\citep{Aakur_2019_CVPR}, Coseg~\citep{wang2021coseg} and ABD~\citep{du2022fast}. 
Due to the demand of clustering methods for a pre-set clusters numbers, we reveal the average number of actions to CTE~\citep{kukleva2019unsupervised}, TW-FINCH~\citep{sarfraz2021temporally}, and ABD~\cite{du2022fast}. Additionally, we also evaluate the naive setting of \textit{Equal Spit}, which equally divides the video into the same number of segments.

\subsection{Evaluation Metrics}
\label{section:metrics}

Owing to the dramatically varying patterns and lengths of the videos as well as the relatively subjective ground-truth annotations for temporal action segmentation, we find current metrics are sometimes too crude to rely on. As the community always pursues a fair and rigorous evaluation, we think it necessary to dive into the details of the metrics.

\noindent\textbf{Frame-level metrics.}
Dense frame-level scores such as Mean of Frames (MoF) or Intersection over Unions (IoU) are the most commonly used metrics for supervised temporal action segmentation~\cite{kuehne2014language,richard2017weakly,chen2020action}, because they accurately measure the distance between the predictions and the ground-truth annotations. In unsupervised settings, however, they become slightly unnatural since the unsupervised algorithms produce untagged neutral segments and we have to rely on the Hungarian algorithm~\citep{kuhn1955hungarian} to artificially build the correspondence between predictions and annotations, which may cause additional biases. Moreover, frame-level metrics are likely to be manipulated by the dominant pattern in the sequence, as illustrated in Figure~\ref{fig:metrics}. This bias can be cross-verified by the quantitative results in Table~\ref{breakfast result}, where even the naive setting of \textit{Equal Split} could achieve relatively decent performance in MoF.

\definecolor{mypurple}{RGB}{216, 131, 255}
\begin{figure}[t]
\centering
\includegraphics[width=0.47\textwidth]{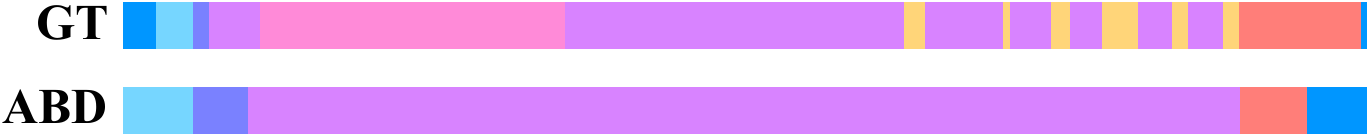}
\vspace{0.5mm}
\caption{\textbf{Example of unreasonable MoF score.} We compare the ground-truth annotations on the clip of \textit{P47\_webcam01\_salad} in Breakfast~\cite{kuehne2014language} and the results generated by ABD~\cite{du2022fast}. While ABD fails to detects more than $80\%$ of the boundaries, it reaches a high MoF of $56.78$, because the MoF is manipulated by the dominant frames standing for \textit{Cut Fruit} marked in \textcolor{mypurple}{purple}.}
\label{fig:metrics}
\end{figure}

\noindent\textbf{Boundary-level metrics.}
Compared with the potentially biased frame-level metrics, we prefer the boundary-level F1 score, which directly measures the discrepancy between the predicted and the annotated boundaries. Nevertheless, the side effect is that we have to manually decide the threshold within which a predicted boundary is accepted as a positive one. To our surprise, the conventional procedure~\citep{wang2021coseg, shou2021generic} is to take $5\%$ of the video duration as the distance threshold (\ie, $15\:\mathrm{seconds}$ for a five-minute video), which we believe is too coarse to distinguish right from wrong.

Thus, we advocate evaluation with F1(\textit{small}) which has a smaller fixed threshold of $2\:\mathrm{seconds}$. With respect to conventions, we also report the \textcolor{gray}{F1(\textit{large})} scores following the traditional $5\%$ threshold, but they are marked in \textcolor{gray}{gray} for distinction.

\noindent\textbf{Human evaluation.}
Though very expensive, the probably most unbiased way of evaluation is always to invite humans to make the judgment, especially when the ground-truth annotations themselves are more or less subjective. Since personal characteristics such as race, gender, \etc, are less likely to impact decisions on action segmentation, we recruit 33 anonymous volunteers from the Internet. Each of them is provided with 20 video clips randomly picked from the Breakfast~\cite{kuehne2014language} dataset with a total length of 40.3 minutes, accompanied with five segmentation results per video in a shuffled order --- one copied from ground-truth, one generated by OTAS, and the other three generated by previous works~\citep{kukleva2019unsupervised, sarfraz2021temporally,du2022fast}. Given no instructions or clues on how to segment (\ie, no granularity reference), the volunteers are asked to rank the five options. Afterward, $(6-\mathrm{rank})$ is considered as the ranking score for the video (\eg, ranking No.2 gets $4\:\mathrm{points}$), and the averaged ranking score is used for evaluation. The average completion time per person is two and a half hours.



 \subsection{Experimental Results}
\label{section:results}

\subsubsection{Quantitative Results}
\label{section:QE}

The quantitative results of OTAS against the baselines in terms of frame-level MoF and boundary-level F1 scores on the three datasets are shown in Table~\ref{breakfast result},~\ref{50Salads result},~\ref{INRIA result}, respectively. F1(\textit{small}) is recommended over MoF because it is in better accord with the requirements for unsupervised action segmentation. \textcolor{gray}{F1(\textit{large})} is also presented but not worth noticing due to its false thresholds.

Equipped with local object attention, OTAS significantly surpasses the previous state-of-the-art method~\cite{du2022fast} by $\mathbf{41\%}$ on average in terms of F1(\textit{small}), and outperforms all the baselines in terms of MoF on Breakfast and 50Salads. 

Note that F1(\textit{small}) reports a performance ranking generally in line with MoF but could better distinguish the competence of different approaches, while \textcolor{gray}{F1(\textit{large})} reports a ranking that is rather inconsistent with its counterparts. These observations could cross-validate our claim on the evaluation metrics.

\begin{table}[t]
\footnotesize
  \centering
  \setlength{\tabcolsep}{3mm}
  \begin{tabular}{@{}lcccc@{}}
    \toprule
    ~&MoF & \textcolor{gray}{F1(\textit{large})} & F1(\textit{small}) \\
    \midrule
    \; Equal Split & 54.06& \textcolor{gray}{33.08} & 14.04   \\
    \; LSTM+AL\citep{Aakur_2019_CVPR}&42.90&\textcolor{gray}{--}&--\\
    \; CTE\citep{kukleva2019unsupervised} &60.50& \textcolor{gray}{35.60} &19.52  \\ 
    \; TW-FINCH\citep{sarfraz2021temporally}& 62.70& \textcolor{gray}{43.35}  & 23.82 \\
    \; CoSeg\citep{wang2021coseg}&53.10& \textcolor{gray}{54.70}  &--\\
    \; ABD\citep{du2022fast}&64.00 &\textcolor{gray}{49.56}&27.93\\
    \; \textbf{OTAS~(ours)}&\textbf{67.90}&\textcolor{gray}{\textbf{62.13}}&\textbf{44.49}\\
    \bottomrule
  \end{tabular}
  \vspace{0.8mm}
  \caption{\textbf{Quantitative results on Breakfast~\cite{kuehne2014language}.} OTAS significantly outperforms the second best by $\mathbf{59\%}$ in terms of F1(\textit{small}), and leads the chart in terms of MoF as well.}
  \label{breakfast result}
\end{table}

\begin{table}[t]
\footnotesize
  \centering
  \setlength{\tabcolsep}{3mm}
  \begin{tabular}{@{}lcccc@{}}
    \toprule
    ~&MoF & \textcolor{gray}{F1(\textit{large})} & F1(\textit{small}) \\
    \midrule
    \textit{Eval-level}\\
     \quad Equal Split & 47.40& \textcolor{gray}{63.97} & 15.92   \\
    \quad CTE\citep{kukleva2019unsupervised}& 53.92&\textcolor{gray}{41.82}&11.80 \\
    \quad LSTM+AL\citep{Aakur_2019_CVPR}&60.60&\textcolor{gray}{--}&--\\
    \quad CoSeg\citep{wang2021coseg}& 64.10 &\textcolor{gray}{71.80}&--\\
    \quad TW-FINCH\citep{sarfraz2021temporally}& 71.10 &\textcolor{gray}{47.15}&22.72\\
    \quad ABD\citep{du2022fast}&71.40&\textcolor{gray}{66.02}&30.50\\
    \quad \textbf{OTAS~(ours)}&\textbf{73.57}&\textcolor{gray}{\textbf{72.72}}&\textbf{49.58}\\
    \midrule
    \textit{Mid-level}\\
    \quad Equal Split & 33.10& \textcolor{gray}{\textbf{78.35}} & 22.88   \\
    \quad CTE\citep{kukleva2019unsupervised}& 43.68 &\textcolor{gray}{41.93}&11.76\\
    \quad TW-FINCH\citep{sarfraz2021temporally}& 66.50 &\textcolor{gray}{70.19}&19.18\\
    \quad ABD\citep{du2022fast}&71.80&\textcolor{gray}{70.80}&40.47\\
    \quad \textbf{OTAS~(ours)}&\textbf{72.42}&\textcolor{gray}{71.07}&\textbf{53.13}\\
    \bottomrule
  \end{tabular}
  \vspace{0.8mm}
  \caption{\textbf{Quantitative results on 50Salads~\cite{stein2013combining}.} OTAS significantly outperforms the second best by $\mathbf{63\%}$ in terms of F1(\textit{small}) on \textit{Eval-level} and by $\mathbf{31\%}$ on \textit{Mid-level}. OTAS also leads the chart in terms of MoF. Notably, Equal Split should beat all the other fancy techniques in terms of the unreasonable \textcolor{gray}{F1(\textit{large})} metrics on \textit{Mid-level}. We attribute the phenomenon to the coincidence that each action of mixing salads takes approximately the same time.}
  \label{50Salads result}
\end{table}

\begin{table}[t]
\footnotesize
  \centering
  \setlength{\tabcolsep}{3mm}
  \begin{tabular}{@{}lcccc@{}}
    \toprule
    ~&MoF & \textcolor{gray}{F1(\textit{large})} & F1(\textit{small}) \\
    \midrule
     \; Equal Split & 30.2& \textcolor{gray}{57.54} & 24.58  \\
     \; CTE\citep{kukleva2019unsupervised}& 39.08 & \textcolor{gray}{\textbf{70.27}}&27.62\\
     \; CoSeg\citep{wang2021coseg}& 47.90&\textcolor{gray}{53.70}&--\\
    \; TW-FINCH\citep{sarfraz2021temporally}& 56.70 & \textcolor{gray}{58.10}&24.27\\
   \; ABD\citep{du2022fast}&\textbf{67.20}&\textcolor{gray}{64.92}&34.18\\
\; \textbf{OTAS~(ours)}&65.71&\textcolor{gray}{66.53}&\textbf{37.28}\\
    \bottomrule
  \end{tabular}
  \vspace{0.8mm}
  \caption{\textbf{Quantitative results on INRIA~\cite{alayrac2016unsupervised}.} OTAS outperforms the second best by $9\%$ in terms of F1(\textit{small}). It achieves comparable performance to ABD~\cite{du2022fast} in terms of MoF. We attribute the degradational performance of OTAS to up to $83\%$ ratio of background frames on the INRIA dataset.}
  \label{INRIA result}
\end{table}

\begin{table}[t]
\footnotesize
  \centering
  \setlength{\tabcolsep}{2mm}
  \begin{tabular}{@{}lc@{}}
    \toprule
    ~&Avg. Ranking Score \\
    \midrule
    \; CTE\citep{kukleva2019unsupervised} & 2.5025\\
    \; TW-FINCH\citep{sarfraz2021temporally} & 2.722\\ 
    \; ABD\citep{du2022fast} & 2.933\\
    \; Ground-Truth&3.167\\
    \; \textbf{OTAS~(ours)}&\textbf{3.331}\\
    \bottomrule
  \end{tabular}
  \vspace{0.8mm}
  \caption{\textbf{User study results on Breakfast.} Humans tend to rate OTAS over all the baselines. Surprisingly, OTAS is even considered to outperform the ground-truth annotations in the dataset.}
  \label{user study result}
\end{table}

\subsubsection{User Study}
\label{section:Us}

\begin{figure}[t]
\centering
\includegraphics[width=0.42\textwidth]{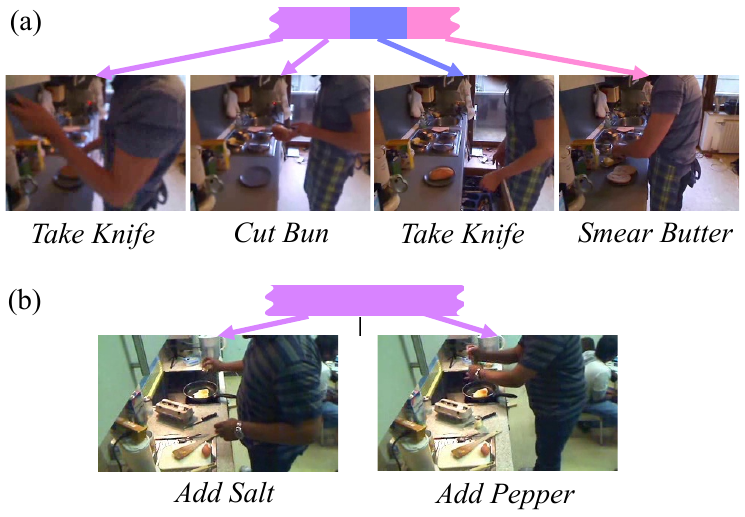}
\caption{\textbf{Examples of common flaws in ground-truth (GT). (a) Inconsistent granularity.} GT distinguishes \textit{Take Knife} and \textit{Smear Butter} but does not distinguish \textit{Take Knife} and \textit{Cut Bun}.
\textbf{(b) Confusion with similar actions.} GT perceives \textit{Add Salt} and \textit{Add Pepper} as one action while users tend to separate them. }
\label{different segmentation}
\end{figure}

As a helpful complement to quantitative metrics, the user study results are listed in Table~\ref{user study result}. Humans tend to rate OTAS over the baselines. They even prefer OTAS to the ground-truth annotations, indicating that human-annotated labels are sometimes counter-intuitive. By carefully examining the dataset, we discover the ground-truth labels are deteriorated by inconsistent segmentation granularity and negligence on fine object details, as illustrated in Figure~\ref{different segmentation}.


Besides, note that the ranking score is largely in line with the F1(\textit{small}) metrics in Table~\ref{breakfast result},~\ref{50Salads result},~\ref{INRIA result}, which once again cross-validates our preference to evaluation metrics.




\subsubsection{Qualitative Results}
\begin{figure*}[t]
\centering
\includegraphics[width=0.96\textwidth]{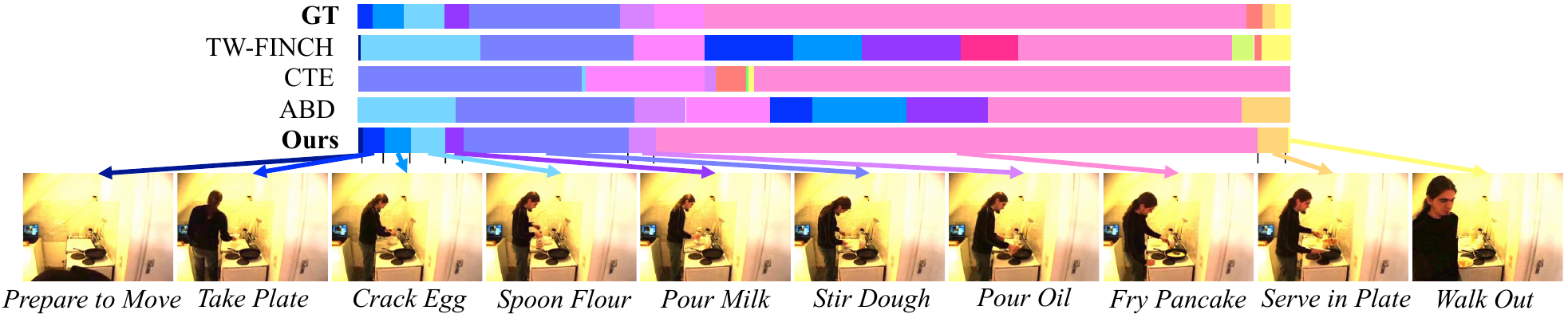}
\caption{\textbf{Case study of the segmentation results of ground-truth (GT), TW-FINCH~\citep{sarfraz2021temporally}, CTE~\citep{kukleva2019unsupervised}, ABD~\citep{du2022fast}, and Ours.} OTAS successfully recognizes consecutive fine actions at the beginning while avoids to over-segment the long action of \textit{Fry Pancake}.}  
\label{qualitative breakfast}
\end{figure*}

The qualitative case study on a challenging video clip with dramatically varying segment lengths is shown in Figure~\ref{qualitative breakfast}.
Thanks to the collaboration between the global and local features, OTAS not only successfully captures the detailed variations easily omitted between consecutive short actions, but also effectively alleviates the over-segmentation problem commonly encountered in a continuous long segment.

\begin{table}[t]
\footnotesize
  \centering
  \setlength{\tabcolsep}{2mm}
  \begin{tabular}{@{}lccc@{}}
    \toprule
    &Feature $\mathrm{(s)}$ & Inference $\mathrm{(s)}$ & Overall $\mathrm{(fps)}$\\
    \midrule
    \; CTE~\citep{kukleva2019unsupervised} & 229.1&217.94&4.5\\
    \; TW-FINCH~\citep{sarfraz2021temporally} & 229.1&0.16&8.7\\ 
    \; ABD~\citep{du2022fast} & 229.1&\textbf{0.02}&8.7\\
    \; \textbf{OTAS~(ours)}&\textbf{5.58}&\textbf{0.02}&\textbf{357.1}\\
    \bottomrule
  \end{tabular}
  \vspace{0.1mm}
  \caption{\textbf{Comparison of the computational cost.} 
 Boundary-based methods (OTAS and ABD~\cite{du2022fast}) are significantly more efficient than clustering-based methods (CTE~\citep{kukleva2019unsupervised} and TW-FINCH~\citep{sarfraz2021temporally}) when inference, while the overall computational cost is bounded by the feature extraction process. OTAS obtains overall efficiency of $357.1~\mathrm{fps}$ on a single NVIDIA GeForce RTX 2080 Ti, indicating its potential for real-time applications.}
  \label{run time}
\end{table}

\subsubsection{Computational cost}
The computational cost is compared in Table~\ref{run time}. Following ABD~\citep{du2022fast}, we report the inference time to segment $2000$ frames with the visual features already prepared. As a boundary-based method, OTAS is as efficient as ABD while evidently more efficient than clustering-based methods.

Additionally, we report the often ignored cost of feature extraction and the more practical overall efficiency in $\mathrm{fps}$, which involves both feature extraction and segmentation inference. Interestingly, we find that the overall expenses are bounded by the feature preparation procedure. By explicitly consuming raw video inputs, OTAS avoids the time-consuming computation of IDT~\cite{wang2013action} features leveraged by the baselines, making it stands out in overall efficiency.

\subsection{Ablation study}
\label{section:ablation}
The ablation study in Table~\ref{Ablation models} demonstrates the effectiveness of the local object attention module. 
A more intuitive comparison between the qualitative behaviors of different feature combinations is shown in Figure~\ref{Ablation models quality}.
More ablation studies can be found in the supplementary materials.

\begin{figure}[t]
\centering
\includegraphics[width=0.47\textwidth]{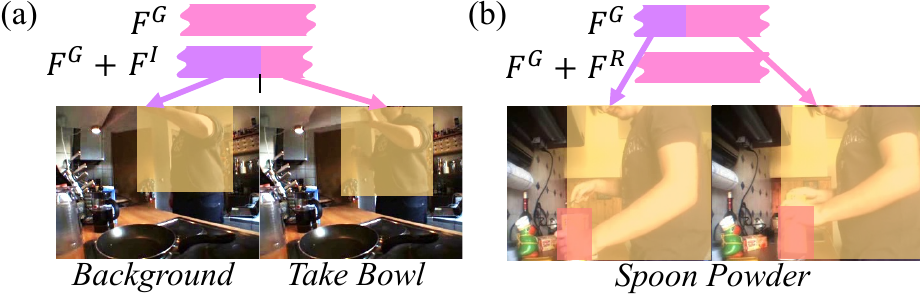}
\vspace{0.1mm}
\caption{\textbf{Effectiveness of local features.} (a) The action of \textit{Take Bowl} occurs in a small interacting region. $\boldsymbol{F}^I$ successfully detects it while $\boldsymbol{F}^G$ 
 alone fails.
(b) \textit{Spoon Powder} is a large-scale action in the near end of the camera perspective, which leads to significant variations in the global feature. $\boldsymbol{F}^R$ avoids over-segmentation with the discovery of invariant object relations.
}
\label{Ablation models quality}
\end{figure}

\begin{table}[t]
\footnotesize
  \centering
  \setlength{\tabcolsep}{0.85mm}
  \begin{tabular}{@{}lcc@{}}
    \toprule
    ~& F1(\textit{small}) & MoF \\
    \midrule
    \; $\boldsymbol{F}^G$&37.46&65.99\\
    \; $\boldsymbol{F}^I$&36.54&65.19\\
    \; $\boldsymbol{F}^R$&37.29&61.33\\
    \; $\boldsymbol{F}^G+\boldsymbol{F}^I$&43.09&65.12\\
    \; $\boldsymbol{F}^G+\boldsymbol{F}^R$&44.11&67.85\\ \; $\boldsymbol{F}^G+\boldsymbol{F}^I+\boldsymbol{F}^R$&\textbf{44.49}&\textbf{67.90}\\
    \bottomrule
  \end{tabular}
  \vspace{0.1mm}
  \caption{\textbf{Ablation study on Breakfast.} The ensemble of global and local features outperforms any sole feature. The ensemble of all three features leads to the best performance.}
  \label{Ablation models}
\end{table}

%% file: 05conclusion.tex
\section{Conclusion}
In this paper, we take local visual features into consideration to tackle the problem unsupervised temporal action segmentation. We discuss the rationality of the evaluation metrics, and more importantly, propose the framework of OTAS that combines global visual features, local interacting features, and object relational features. OTAS achieves state-of-the-art performance on quantitative metrics, user studies, and computational cost.